\documentclass[letterpaper, 10 pt, conference]{ieeeconf}  
\def\IEEEtitletopspace{18pt}

\IEEEoverridecommandlockouts                              

\usepackage{graphicx} 
\usepackage{float} 
\usepackage{cite}
\usepackage{booktabs}
\usepackage{bbding}
\usepackage{color}
\usepackage{threeparttable}
\usepackage{multirow}
\usepackage{amssymb}
\usepackage{amsmath}

\usepackage[section]{placeins}
\usepackage[colorlinks,linkcolor=red,anchorcolor=blue,citecolor=green]{hyperref}
\title{\LARGE \bf
A Fast and Map-Free Model for Trajectory Prediction in Traffics
}

\author{
Junhong Xiang$^{1,{\dagger}}$, Jingmin Zhang$^{2,{\dagger}}$ and Zhixiong Nan$^{1,*}$
\thanks{This work is supported by National Key Research and Development Program of China (No.2020AAA0108100), National Natural Science Foundation of China (NO.62006180) and Joint Fund of Ministry of Education of China for Equipment Pre-Research(No.8091B032127).}
\thanks{$^{1}$College of Computer Science, Chongqing University, Chongqing 400044, China.}%
\thanks{$^{2}$No.208 Research Institute of China Ordnance Industries, Beijing 102202, China.}%
\thanks{$\dagger$ Junhong Xiang and Jingmin Zhang are co-first authors.}%
\thanks{${*}$ Zhixiong Nan is the corresponding author.}
}

\begin{document}

\maketitle

\begin{abstract}
To handle the two shortcomings of existing methods, (i)
nearly all models rely on high-definition (HD) maps, yet the map information is not always available in real traffic scenes and HD map-building is expensive and time-consuming and (ii) existing models usually focus on improving prediction accuracy at the expense of reducing computing efficiency, yet the efficiency is crucial for various real applications, this paper proposes an efficient trajectory prediction model that is not dependent on traffic maps. The core idea of our model is encoding single-agent's spatial-temporal information in the first stage and exploring multi-agents' spatial-temporal interactions in the second stage. By comprehensively utilizing attention mechanism, LSTM, graph convolution network and temporal transformer in the two stages, our model is able to learn rich dynamic and interaction information of all agents. 
Our model achieves the highest performance when comparing with existing map-free methods and also exceeds most map-based state-of-the-art methods on the Argoverse dataset. In addition, our model also exhibits a faster inference speed than the baseline methods.


\end{abstract}

\section{INTRODUCTION}
In the field of autonomous driving, trajectory prediction is an important topic which targets to predict the intentions of traffic agents, enabling an autonomous agent to make more smart planning\cite{jian2021global,zhang2021hierarchical}. Therefore, efficient and accurate trajectory prediction in complex traffic scenarios has essential research significance. However, in traffic scenarios, each agent's behavior\cite{zhang2020driving} (e.g, changing lanes, accelerating, turning) is random and dynamic in both temporal and spatial dimensions and the interactions between multiple agents are complex, thus the prediction of agent trajectory is challenging.

In early days, a kind of approach models the interaction information between traffic agents by rasterizing the driving scene into a bird's eye view image so that the environmental information could be efficiently processed by a convolutional neural network (CNN) \cite{cui_multimodal_2019,chai_multipath_2019,zhao_multi-agent_2019}. As rasterization suffers from quantization errors, high computational costs of processing rendered images and a restricted field of view, recent works adopt the contextual encoding approach based on vector data obtained directly from HD maps \cite{zhou_hivt_2022,kim2021lapred,gao_vectornet_2020,liang2020learning}. For example, VectorNet\cite{gao_vectornet_2020} treats the historical trajectories of lanes and agents on a map as a set of folds and models them as a fully connected interaction graph; LaneGCN\cite{liang2020learning} proposes to organize the lanes into a lane graph by taking spatial connectivity into account, and then uses a graph convolutional network (GCN) to encode the topology of the map.
 \begin{figure}[!t]
 	\centering
 	\includegraphics[width=0.48\textwidth]{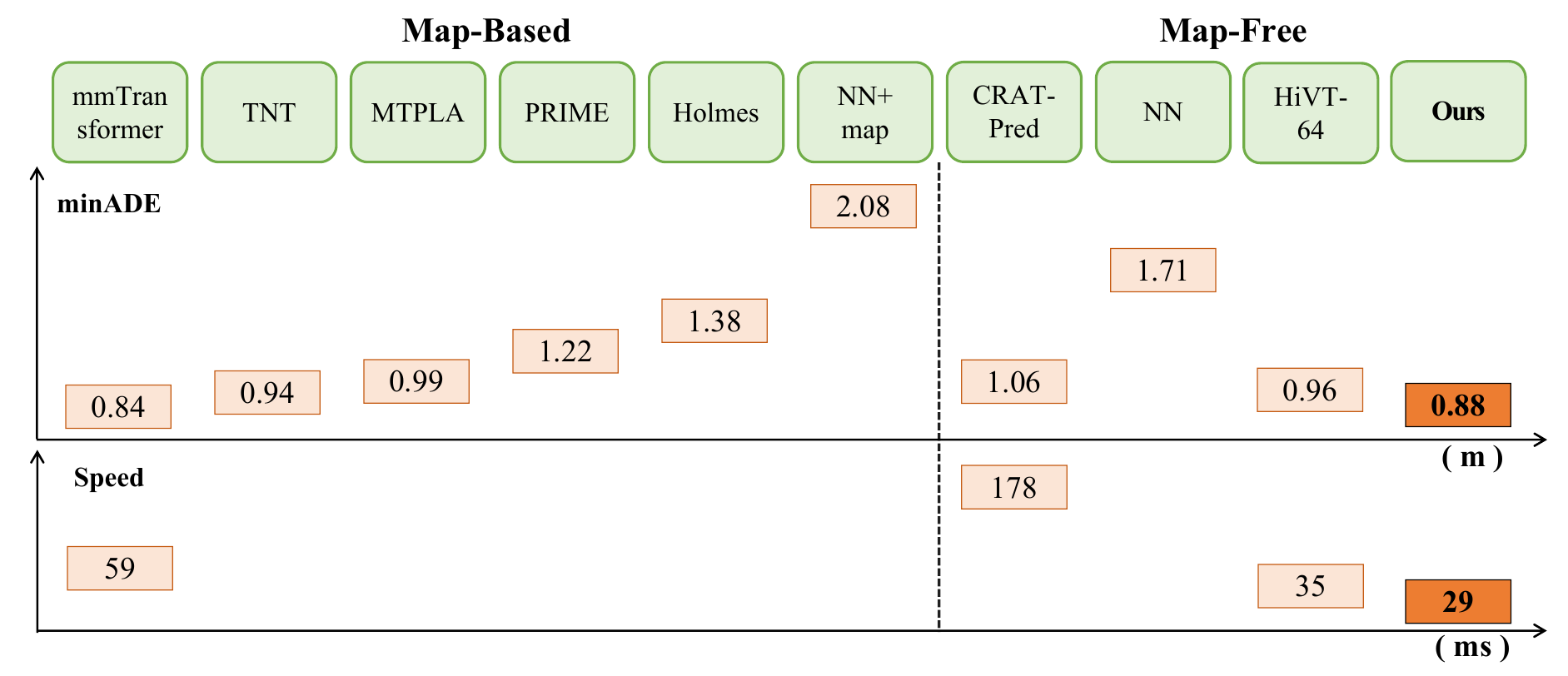}
 	\caption{The inference speed and the prediction performance of the models on the Argoverse dataset.}
 	\label{introduction}
   \vspace{-0.5cm}
 \end{figure}
These vehicle trajectory prediction models rely on map information, which is realistically not always available in open environments.Although, there are many models that deal with trajectory prediction without using map information \cite{cheng2021amenet,schmidt2022crat}, the prediction accuracy is not yet comparable to the models based on the map information. Therefore, it is of great significance to study a map-free trajectory prediction model that presents competitive performance with map-based models.

Prevailing models tend to prioritize enhancing predictive accuracy while compromising computational efficiency, as shown in the Fig. \ref{introduction}. However, the inference speed is of paramount importance for diverse practical applications. Recently, Transformer networks have performed well in the field of trajectory prediction. Huang et al.\cite{huang2022multi} proposes a multimodal attention transformer encoder to generate multimodal trajectories. Zhou et al.\cite{zhou_hivt_2022} employs a time-transformer encoder similar to BERT to capture temporal information in local regions. Liu et al.\cite{liu_multimodal_2021} uses a stacked transformer-based network structure to integrate environmental contextual information in a hierarchical manner.

Since the attention mechanism is used in the transformer network, the computational complexity of the attention mechanism is quadratic to the length of the input sequence. For traffic agents with input patterns of temporal ($T$) and spatial ($S$) dimensions, the computational cost reaches $\mathcal{O}\left(S^{2} \times T^{2}\right)$, which severely affects the inference efficiency of the trajectory prediction model and hinders the applications in real-time scenarios. Although many approaches simplify the Transformer model architecture by simply using only the Transformer encoder module, this does not fundamentally solve the problem. To alleviate the computational cost strain on the trajectory prediction model, we consider applying attention in both the temporal and spatial dimensions separately, which reduces the cost of the attention mechanism network from $\mathcal{O}\left(S^{2} \times T^{2}\right)$ to $\mathcal{O}\left(S^{2}\right)+\mathcal{O}\left(T^{2}\right)$.

Based on the above observations, we propose a fast agent trajectory prediction model that does not rely on map information.
Our model consists of two stages. In the first stage, an attention mechanism is firstly applied on the history trajectories of each agent to extract spatial features ($S_A$), and a ``LSTM+attention" structure is designed to extract the temporal features ($T_A$). To explore the context relations of multiple agents, $S_A$ and $T_A$ are then fused and further processed in the second stage. In the second stage, a graph convolutional network (GCN) based interaction module is used to learn inter-agent interaction in the spatial dimension, and a temporal Transformer module is utilized to capture inter-agent interaction in the temporal dimension. Finally, a decoder is employed to generate multimodal trajectories for each agent.

Our contributions are as follows:
(1) A map-free method for trajectory prediction which outperforms several state-of-the-art map-based models with less information. 
(2) Novel spatial and temporal feature decoupling on both the single agent and the inter-agent level, which is validated to be effective through our results.
(3) Faster inference speeds compared to both state-of-the-art map-free and map-based models.

\section{RELATED WORKS}

\textbf{Sequence modelling} is one main strategy for the trajectory prediction, which determines whether the model can effectively extract historical trajectory sequence features. The ability of RNNs to store time-step information has made RNNs the model of choice for trajectory and motion prediction.
Long Short Term Memory (LSTM) can solve the gradient explosion problem and therefore many studies have used LSTM to model trajectory sequence features. For example, Altche et al.\cite{altche_lstm_2017} uses LSTM to extract features of vehicles and feed their final hidden layer to the output layer to predict vehicle trajectories, while Park et al.\cite{park_sequence--sequence_2018} uses an LSTM-based encoder to analyze the patterns of past trajectories and another decoder to generate sequences of future trajectories. With the introduction of the Transformer, it has been widely used to model sequence features in trajectory prediction tasks due to its ability to capture long-range dependencies and filter out high-value information from large amounts of information. For example, Messaoud et al.\cite{messaoud_attention_2020,huang_multi-modal_2021} uses a multi-headed attention mechanism to extract trajectory features, where Huang et al.\cite{huang_multi-modal_2021} also considers the interactions between agent organisations, and Huang et al.\cite{ huang2022multi}, on the other hand, proposes a multimodal attentional transformer encoder to generate multimodal trajectories. Such methods can severely affect the real-time predictive capability of the model because of the computational pressure of the attention mechanism. Therefore, we designed a single agent encoder module with a multidimensional attention mechanism to model the time-series and spatial-series features of a single agent, which improves the inference speed of the model.

\textbf{Interaction modelling} is the basis for capturing interaction information from a scene map or a dynamic agent. The most common method for encoding the interaction information between the map and traffic participants, taking into account drivable areas and HD maps, is to rasterize the driving scene into a bird's eye view image. Such a representation of the environment can be efficiently processed by convolutional neural networks (CNNs), which have been used to good effect in many motion prediction works \cite{cui_multimodal_2019,chai_multipath_2019,zhao_multi-agent_2019,yang2020traffic}. Cui et al.\cite{cui_multimodal_2019 } splices raster images with historical trajectory information after CNN processing and passes them into a multilayer fully connected layer to complete the trajectory prediction task. CoverNet\cite{phan-minh_covernet_2019} forms the trajectory multimodal prediction by classifying the vehicle state (speed, acceleration and yaw rate) and raster images. As rasterisation suffers from quantization errors and a restricted field of view, many studies construct graph structures to simulate social interactions based on vector data obtained from HD maps, and such methods are popular for their efficient sparse coding and ability to capture complex structural information. CRAT-Pred\cite{schmidt2022crat} applies a graph convolution method originating from the field of material science to vehicle prediction, allowing to efficiently leverage edge features, and combines it with multi-head self-attention. DSP\cite{zhang2022trajectory} proposes a graph-based trajectory prediction network that encodes static and dynamic interaction environments in a hierarchical manner. VectorNet\cite{gao_vectornet_2020} models the interactions between lanes and trajectory folds using graph neural networks. LaneGCN\cite{liang_learning_2020} proposes to organize the lanes on a map into a lane graph that takes spatial connectivity into account, and then use graph convolutional networks (GCN) to encode the topology of the map for more effective context fusion.
Most of the state-of-the-art methods listed above are designed based on HD maps, limiting their use in scenarios where map information is not available. Our model is therefore specifically designed for map-free trajectory prediction, combining GNN and Transformer to design a multi-agent interaction module that exploits as much information as possible about the interaction relations, obtaining a scenario-independent and high-performance trajectory prediction model.

\section{Method}
\begin{figure*}[h]
 	\centering
 	\includegraphics[width=1\textwidth]{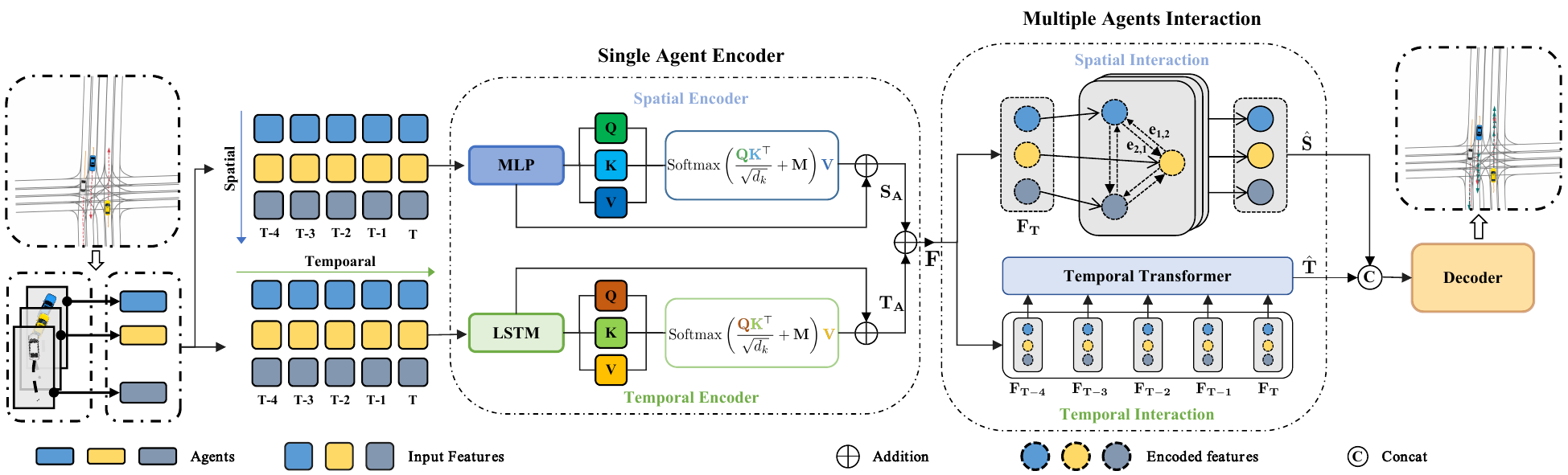}
 	\caption{
  Our model consists of two stages. Given the history trajectories of multiple agents as the input, in the first stage, an attention mechanism is firstly applied on the input feature of each agent to extract spatial features ($S_A$), and ``LSTM+attention" network extracts the temporal features ($T_A$) at the same time. To explore the context relations of multiple agents, $S_A$ and $T_A$ are then fused and further processed in the second stage, where a graph convolutional network (GCN) based spatial interaction module is used to learn inter-agent interaction in the spatial dimension, and a temporal Transformer module is utilized to capture inter-agent interaction in the temporal dimension. Finally, the spatial interaction representation ($\hat{\mathbf{S}}$) and the temporal interaction representation ($\hat{\mathbf{T}}$) are fed into a decoder that generates multimodal trajectories for each agent.
  }
 	\label{overview}
  \vspace{-0.3cm}
\end{figure*}

An overview of our proposed trajectory prediction model is presented in Fig. \ref{overview}. In the following, we first formulate the problem, and then detail the proposed model. 
\subsection{Problem Formulation}
The goal of multimodal trajectory prediction is to predict multiple possible future trajectories of a target agent based on the state features of the target agent and its surrounding agents. 
In a scenario with $N$ agents, the state features are defined as:
\begin{equation}
    \boldsymbol{X}=\left\{\boldsymbol{\tau}_{i}^{t} \mid i \in 1, \ldots, N ; t \in-T_{h}+1, \ldots, 0\right\}
\end{equation}
where $T_h$ denotes the historical time horizon, at each time step $t$, the coordinates of vehicle $i$ are denoted as $\boldsymbol{\tau}_{i}^{t} = \{\mathbf{x}_{i}^{t},\mathbf{y}_{i}^{t}\}$.
Based on the available feature in this scenario, the multimodal trajectory prediction task can be represented as predicting
\begin{equation}
\hspace{-2mm}
    \boldsymbol{Y}=\left\{\hat{\boldsymbol{\tau}}_{i,k}^{l} \mid i \in 1, \ldots, N ; l \in 1, \ldots, T_{f};k \in 1, \ldots, K \right\}
\end{equation}
with $T_f$ denotes the prediction time horizon, $k$ represents the number of patterns of multiple predicted trajectories and $\hat{\boldsymbol{\tau}}_{i,k}^{t} = \{\mathbf{x}_{i,k}^{t},\mathbf{y}_{i,k}^{t}\}$.

\subsection{Single Agent Encoder}
\subsubsection{Feature Extraction}
Trajectory prediction heavily relies on effective feature extraction as a crucial step. First, to avoid input features being influenced by the position of the target agent $i$, we take the position of the target agent at its last observed time step and use the latest trajectory segment $\boldsymbol{\tau}_{i}^{0} - \boldsymbol{\tau}_{i}^{-1}$ of agent $i$ as a reference vector for the local region. Rotating all local vectors according to the direction $\theta_{i} $ of the reference vector achieves that all vectors are normalised. After that,
to mitigate the variability of coordinates across scenarios, inspired by the works in \cite{cheng2021exploring,zhou_hivt_2022}, the relative position rather than the absolute position is used to construct state information as follow:
\begin{equation}
\hspace{-2mm}
\boldsymbol{R}_{i}=\left\{ [\Delta \boldsymbol{\tau}_{i}^{t},f_{i}^{t}] \mid i \in 1, \ldots, N ; t \in-T_{h}+1, \ldots, 0\right\}
\end{equation}
where $\Delta \boldsymbol{\tau}_{i}^{t} = \{\Delta x_{i}^{t},\Delta y_{i}^{t}\}$ is the offset from $t-1$ to the next time step, $\Delta \boldsymbol{\tau}_{i}^{-T_{h}+1} = \{0,0\}$. We also only consider vehicles that are observable at $t = 0$ and handle vehicles that are not observed over the historical time horizon $T_{h}$ by concatenating a binary flag $f_{i}^{t}$.

\subsubsection{Spatial \& Temporal Encoder}Information about the trajectory of each agent is encoded in spatial and temporal dimensions by the Spatial \& Temporal Encoder. The spatial dimension learns information about the spatial characteristics of each individual agent at each historical time step, and the temporal dimension aims to learn information about the dynamics of each agent over the historical time horizon. The module takes as input the state information $\boldsymbol{R}_{i}$, which is then passed to the LSTM and the position-wise multilayer perceptron (MLP) acting on the temporal and spatial dimensions, respectively. 

The LSTM is formulated as
\begin{equation}
\mathbf{H}_{i}^{t}=\operatorname{LSTM}\left(\mathbf{H}_{i}^{t-1}, \mathbf{R}_{i}^{t}, \mathbf{W}_{\mathrm{enc}}, \mathbf{b}_{ \text {enc }}\right)
\end{equation}
where $\mathbf{W}_{\mathrm{enc}}$ and $\mathbf{b}_{ \text {enc }}$ are learnable parameters. We use the two-layer LSTM structure with shared weights for all agents, the hidden state $\mathbf{H}_{i}^{t}$ is a vector of size 128. The MLP contains two fully connected layers and the output dimension is consistent with the hidden state vector size of the LSTM.

After LSTM and MLP, updated temporal state information $\mathbf{H}^{t}$ holds about the temporal features of individual agents. To focus attention on salient temporal and spatial features, a multi-head attention layer is applied to each dimension individually, learning which time steps and agents should be given more attention.
For the temporal dimension specifically, the attention mechanism is applied to the temporal feature matrix $\mathbf{H}$. Each head $h \in 1, \ldots, N_{h}$ defined as

\begin{equation}
\operatorname{head}_{h}=\operatorname{softmax}\left(\frac{\mathbf{V}_{Q_{h}}^{\left(H\right)} \mathbf{V}_{K_{h}}^{\left(H\right) T} }{\sqrt{d}}\right) \mathbf{V}_{V_{h}}^{\left(H\right)}
\end{equation}

$\mathbf{V}_{Q_{h}}^{\left(H\right)}$, $\mathbf{V}_{K_{h}}^{\left(H\right)}$ and $\mathbf{V}_{V_{h}}^{\left(H\right)}$ are linear projections of the head $h$ onto the matrix $\mathbf{H}$, and $d$ is a normalization factor. Finally, the temporal features $\mathbf{T_A}$ after applying the attention mechanism in the time dimension is computed by
\begin{equation}
\mathbf{T_A}=Concat(\operatorname{head}_{1},\ldots, \operatorname{head}_{N_{h}}) \mathbf{W}+\mathbf{b}
\end{equation}

In our implementation, eight attention heads are used, i.e. $N_{h}=8$ and $d=\frac{hiden\_size}{N_{h}} $. $\mathbf{W}$ denotes the coefficient matrix and $\mathbf{b}$ is the bias. Note that the attention mechanism for the spatial dimension is the same as for the temporal dimension, except that it is applied to the spatial dimension of each agent. Therefore, the computation of spatial features $\mathbf{S_A}$ is not shown.

\subsection{Multiple Agents Interaction}
\subsubsection{Spatial Interaction}
While the encoder module has initially extracted information about the spatial features of individual agents, the spatial interaction features between multiple agents are also important. Therefore, to further model the interactions between all concurrent agents present in the scenario, a spatial interaction module was constructed based on GCN. To mitigate the variability between coordinates at different historical time steps, the input trajectory features are translational transformed relative position information, but this translational invariant representation loses the relative position information between vehicles. Therefore, in the spatial interaction module, the difference between the coordinate frames of agent $i$ and agent $j$ can be parameterised by $\boldsymbol{\tau}_{j}^{0} - \boldsymbol{\tau}_{i}^{0}$ and $\Delta \theta_{i j}$, where $\Delta \theta_{i j}$ denotes $\theta_{j} - \theta_{i}$. MLP is used to obtain the pairwise embedding $e_{ij}$ when the GCN performs the message passing from agent j to i:
\begin{equation}
 \mathbf{e}_{i j}=\mathbf{MLP} \left(\left[(\boldsymbol{\tau}_{j}^{0} - \boldsymbol{\tau}_{i}^{0}), \cos \left(\Delta \boldsymbol{\theta}_{i j}\right), \sin \left(\Delta \boldsymbol{\theta}_{i j}\right)\right]\right)   
\end{equation}

The embedding of the target agent is then converted to a query vector, and the embedding of the neighbouring agents and the paired embedding $e_{ij}$ are used to compute the key and value vectors.
\begin{equation}
\begin{aligned}
{\mathbf{q}}_{i} & =\mathbf{W}^{Q^{\text {global }}} {(S_A ^i+T_A ^i)}, \\
{\mathbf{k}}_{i j} & =\mathbf{W}^{K^{\text {global }}}\left[(S_A ^j+T_A ^j), \mathbf{e}_{i j} \right], \\\
{\mathbf{v}}_{i j} & =\mathbf{W}^{V^{\text {global }}}\left[(S_A ^j+T_A ^j), \mathbf{e}_{i j}\right],
\end{aligned}
\end{equation}
where $\mathbf{W}^{Q^{\text {global }}}$, $\mathbf{W}^{K^{\text {global }}}$, and $\mathbf{W}^{V^{\text {global }}}$ are linear projection learnable matrices, and the resulting query, key and value vectors are fed into the Scaled dot product multi-head attention blocks:
\begin{equation}
\boldsymbol{\alpha}_{i}=\operatorname{softmax}\left(\frac{\mathbf{q}_{i}^{\top}}{\sqrt{d_{k}}} \cdot\left[\left\{\mathbf{k}_{i j}\right\}_{j \in \mathcal{N}_{i}}\right]\right), \\
\end{equation}
\begin{equation}
\mathbf{m}_{i}=\sum_{j \in \mathcal{N}_{i}} \boldsymbol{\alpha}_{i j} \mathbf{v}_{i j}, \\
\end{equation}
\begin{equation}
\mathbf{g}_{i}=\operatorname{sigmoid}\left(\mathbf{W}^{\text {gate }}\left[(S_A ^i+T_A ^i), \mathbf{m}_{i}\right]\right), \\
\end{equation}
\begin{equation}
{\hat{\mathbf{S}}_{i}}=\mathbf{g}_{i} \odot \mathbf{W}^{\text {self }} (S_A ^i+T_A ^i)+\left(1-\mathbf{g}_{i}\right) \odot \mathbf{m}_{i},
\label{S1} 
\end{equation}
where $\mathcal{N}_{i}$ is the set of agent $i$ neighbours, and $\mathbf{W}^{\text {gate }}$ and $\mathbf{W}^{\text {self }}$ are learnable matrices, $\odot$ denotes element-wise product. We use a gating function to fuse the interaction feature $\mathbf{m}_{i}$ with the spatil-temporal feature $S_A ^i+T_A ^i$ of the target agent, enabling the block to have more control over feature updates.
After the attention module, a MLP module is employed on all agent nodes to generate a spatial interaction representation denoted as $\hat{\mathbf{S}}$.

\subsubsection{Temporal Interaction}

In order to capture the temporal relation of multiple agents in traffic congestion, a temporal interaction module has been designed to compensate for the lack of a single agent temporal encoder.
The output of this module is the temporal interaction representation $\hat{\mathbf{T}}$, the inputs are $\mathbf{T_A}$ and $\mathbf{S_A}$, which are fed into the temporal interaction module as follows:
\begin{equation}
\begin{aligned}
\mathbf{Q}=( \mathbf{S_A} + \mathbf{T_A} ) \mathbf{W}^{Q^{\text {time }}}, \\ \mathbf{K}=( \mathbf{S_A} + \mathbf{T_A} ) \mathbf{W}^{K^{\text {time }}}, \\\mathbf{V}=( \mathbf{S_A} + \mathbf{T_A} ) \mathbf{W}^{V^{\text {time }}},
\end{aligned}
\end{equation}

\begin{equation}
\hat{\mathbf{T}}=\operatorname{softmax}\left(\frac{\mathbf{Q} \mathbf{K}^{\top}}{\sqrt{d_{k}}}+\mathbf{M}\right) \mathbf{V},
\label{T1}
\end{equation}
where $\mathbf{W}^{Q^{\text {time }}}, \mathbf{W}^{K^{\text {time }}}, \mathbf{W}^{V^{\text {time }}} $ is the learnable matrix. In contrast to traditional attention mechanisms, we apply a padding mask $\mathbf{M} \in \mathbb{R}^{N \times T_h}$  to fill in the time steps where multiple agents are invalid when computing attention weights, forcing attention to focus only on temporal features that are valid between multiple agents.
\subsection{Multimodal Decoder}

In the realm of autonomous driving, the future trajectories of traffic agents exhibit inherent multimodality. To address this, we employ the Laplace Mixture Density Network (MDN) decoder\cite{cheng2022gatraj,zhou_hivt_2022} for generating the future multimodal trajectories of the agents. The decoder takes the spatial interaction representation $\hat{\mathbf{S}}$ Eq. $\ref{S1}$ and the temporal interaction representation $\hat{\mathbf{T}}$ Eq. $\ref{T1}$ as inputs and produces a set of predicted distributions $\sum_{k=1}^{K} \pi_{k} \ Laplace (\mu, b)$. Here, $\pi_{k}$ represents the probabilities associated with different modalities, and $\sum_{k=1}^{K} \pi_{k}=1$. The variables $\mu$ and $b$ signify the future position of the agent and its corresponding uncertainty parameters, respectively.

The output from the regression head has a shape of $[K, N, T_f, 4]$, where $K$ represents the total number of trajectory modalities, $N$ signifies the number of agents present in the scene, and $T_f$ indicates the predicted time horizon. Additionally, we employ another MLP and a softmax function to generate the probabilities associated with different modes for each agent. These probabilities are organized in the shape of $[N, K]$.

\subsection{Training}
The total loss of our designed trajectory prediction model is divided into two components, classification loss $\mathcal{L}_{\text {cls}}$ and regression loss $\mathcal{L}_{\text {reg}}$, both of which are equally weighted. Since we only optimise the best pattern $k^*$ among $K$ predictions during training, the error between the actual ground position of the target agent $i$ and the position predicted by the model is first calculated and the trajectory with the smallest error is selected by Eq.$~\ref{min_k}$.

\begin{equation}
k^{*}=\underset{k \in K}{\arg \min }\left\| \hat{\boldsymbol{\tau}}_{i,k}-\boldsymbol{\tau}_{i}\right\|^{2} \\\\
\label{min_k}
\end{equation}
Then, the negative log likelihood of the Laplace distribution is used as the regression loss and the cross-entropy loss as the classification loss for pattern optimisation.

\begin{equation}
\mathcal{L}=\mathcal{L}_{\text {reg }}+\mathcal{L}_{\text {cls }} ,
\end{equation}
\begin{equation}
\mathcal{L}_{\text {reg }}=\frac{1}{NT_f} \sum_{i=1}^{N} \sum_{t=T+1}^{T+T_f}-\log P\left(\mathbf{\tau}_{i}^{t} \mid \hat{\mathbf{\tau}}_{i , k^{*}}^{t}, \mathbf{b}_{i, k^{*}}^{t}\right) ,\\\
\end{equation}
\begin{equation}
\mathcal{L}_{\text {cls }}=\frac{1}{N} \sum_{i=1}^{N} \sum_{k=1}^{K}-\pi_{i, k} \log \left(\hat{\pi}_{i, k}\right) , 
\end{equation}
where $\mathrm{P}(\cdot \mid \cdot)$ is the probability density function of the Laplace distribution, $\hat{\pi}_{i, k}$ is the predicted probability and $\pi_{i, k}$ is our target probability which is a soft displacement error. 
\begin{table*}[ht]
\hspace{-0.1cm}
	\begin{minipage}[t]{0.48\textwidth}
            \centering
            \caption{Performance of several prediction methods evaluated on Argoverse test set}
	\begin{center} 
	\renewcommand{\arraystretch}{1.5}
	\setlength{\tabcolsep}{4pt}{
        \scalebox{0.75}{
	\begin{tabular}{cccccccc} 
		\toprule[1pt] 
		Methods&Models & Conference& minADE $\downarrow$ & minFDE $\downarrow$ & MR(\%) $\downarrow$ & \\
		\cmidrule(lr){1-7}
		\multirow{4}*{Map-free Model}
		~ & NN\cite{chang2019argoverse}& CVPR 2019 & 1.71 & 3.28&53.70 \\
		~ &HiVT-64\cite{zhou_hivt_2022} \textcolor{red}{$\dagger$} &CVPR 2022 & 0.96 &1.69 &23.29  \\
		~ &CRAT-Pred\cite{schmidt2022crat}&ICRA 2022& 1.06 & 1.90 &26.00 \\
		\cmidrule(lr){1-7}
		\multirow{7.4}*{Map-based Model}
		~ & NN+map\cite{chang2019argoverse}& CVPR 2019 & 2.08 &4.02 &58.00 \\ 
            ~& MTPLA\cite{luo2020probabilistic} & IROS 2020 &0.99 & 1.71 & 19.00 \\
		~& Holmes\cite{huang2020diversitygan} & ICRA 2020 &1.38 & 2.66 & 42.00 \\
		~ &TNT\cite{zhao2021tnt}&PMLR 2021 & 0.94 & 1.54&13.28 \\
		~ &PRIME\cite{song2022learning}&CORL 2021&  1.22 & 1.56&\textbf{11.50} \\
		~ &mmTransformer\cite{liu_multimodal_2021}&CVPR 2021& 0.84 &1.32 &15.22 \\
            ~ &HiVT-64\cite{zhou_hivt_2022} &CVPR 2022&\textbf{0.83} &\textbf{1.31} &15.32 \\
		\cmidrule(lr){1-7}
		~ &\textbf{Ours (map free)}& --- & 0.93 & 1.59&21.39 \\
		\bottomrule[1pt] 
	\end{tabular} 
	} 
        }
	\end{center}
        \begin{tablenotes}
        \footnotesize
        \item  \textcolor{red}{$\dagger$} indicates a map-based model that can perform map-free predictions by excluding the map module.
        \end{tablenotes}
	\label{baselines1} 
        \end{minipage}
        \hfill
	\begin{minipage}[t]{0.48\textwidth}
		\centering

        \caption{Performance of several prediction methods evaluated on Argoverse validation set}
	\begin{center} 
	\renewcommand{\arraystretch}{1.5}
	\setlength{\tabcolsep}{4pt}{
        \scalebox{0.75}{
	\begin{tabular}{cccccccc} 
		\toprule[1pt] 
		Methods & Models& Conference& minADE $\downarrow$ & minFDE $\downarrow$ & MR(\%) $\downarrow$ & \\
		\cmidrule(lr){1-7}
            \multirow{4}*{Map-free Model}
		~ & LaneGCN\cite{liang_learning_2020}\textcolor{red}{$\dagger$}& ECCV 2020 &0.79 &1.29 & --- \\
            ~ & Tpcn\cite{ye_tpcn_2021} \textcolor{red}{$\dagger$}& CVPR 2021&0.82 &1.32& 15.00 \\
            ~ &HiVT-64\cite{zhou_hivt_2022} \textcolor{red}{$\dagger$} &CVPR 2022&0.76&1.24&13.86  \\
		~ &CRAT-Pred\cite{schmidt2022crat} & ICRA 2022 &0.85 &1.44 &17.00 \\
		\cmidrule(lr){1-7}
		\multirow{7.4}*{Map-based Model}
            ~& DATF\cite{park2020diverse} & ECCV 2020  &0.92 & 1.52&--- \\
		~& MTPLA\cite{luo2020probabilistic} & IROS 2020  &1.05 & 2.06&--- \\
		~ & DESIRE\cite{lee2017desire}& CVPR 2021  &1.09 & 1.89&--- \\
		~& LaPred \cite{kim2021lapred} & CVPR 2021&0.71& 1.44&--- \\
		~& mmTransformer\cite{liu_multimodal_2021} & CVPR 2021& 0.72&1.21& \textbf{9.20}\\
            ~ &HiVT-64\cite{zhou_hivt_2022} &CVPR 2022&\textbf{0.69} &\textbf{1.04} &10.00  \\
		\cmidrule(lr){1-7}
		~ &\textbf{Ours (map free)}& --- & 0.74 & 1.18 & 11.69 \\
		\bottomrule[1pt] 
	\end{tabular} 
	} 
        }
	\end{center}
        \begin{tablenotes}
        \footnotesize
        \item  \textcolor{red}{$\dagger$} indicates a map-based model that can perform map-free predictions by excluding the map module.
        \end{tablenotes}
	\label{baselines val} 
 		\end{minipage}
    \vspace{-0.3cm}
    \hspace{0.2cm}
\end{table*}
\section{EXPERIMENTS}
\subsection{Experimental Setup}
\subsubsection{Dataset} 


The Argoverse\cite{chang2019argoverse} dataset consists of driving sequences from 324,557 scenes collected in Miami and Pittsburgh, including 205,942 training sequences, 39,472 validation sequences and 78,143 test sequences. For each scene, the trajectories of multiple vehicles sampled at 10HZ are provided. The trajectory prediction task is to predict the future trajectory (3 seconds) of one target agent while considering the past trajectories (2 seconds) of all vehicles in the sequence. Therefore, the length of the sequence in the training and validation set is 5 seconds, while the sequence in the test set contains only the first 2 seconds of motion.

\subsubsection{Evaluation Metrics}
In this section, we use three constant evaluation metrics to evaluate our model, minimum average displacement error (minADE), minimum final displacement error (minFDE) and miss rate (MR) for multimodal (K=6).

\textbf{Minimum Average Displacement Error:} the minimum average displacement error between the predicted trajectory and the true trajectory of the target vehicle over $K$ predictions.
\begin{equation}
\mathrm{minADE}_{K}=\frac{1}{N} \frac{1}{T_f} \mathrm{min}_{k=1}^{K} \sum_{i=1}^{N} \sum_{t=T+1}^{T+T_f}\left\|\hat{\tau}_{i, k}^{t}-\tau_{i}^{t}\right\|_{2}
\label{ade}
\end{equation}

\textbf{Minimum Final Displacement Error:} the minimum final displacement error between the predicted endpoint and the true endpoint out of k predictions.
\begin{equation}
\mathrm{minFDE}_{K}=\frac{1}{N} \mathrm{min}_{k=1}^{K} \sum_{i=1}^{N}\left\|\hat{\tau}_{i, k}^{T+T_f}-\tau_{i}^{T+T_f}\right\|_{2}
\label{fde}
\end{equation}

For Eq.$~\ref{ade}$ and Eq.$~\ref{fde}$, $N$ is the total number of agents. $ K $ denotes that we generate $K $ predictions for each agent and report the best one measured by ADE and FDE, respectively.

\textbf{Miss Rate:} the ratio of sequences where a predicted endpoint is less than 2 meters from the true endpoint.

\subsubsection{Implementation Details}
Our model was trained on an RTX 3090 GPU using the AdamW optimizer with hidden layer size, batch size, initial learning rate, weight decay and dropout rate set to $128$, $32$, $3 \times 10^{-4}$, $1 \times 10^{-4}$ and $0.1$, respectively. The learning rate was decayed using the cosine annealing scheduler. Our model consists of two modules, single agent encoder and multiple agents interaction, the latter consisting of three spatial interaction layers and four temporal interaction layers. The number of layers for all LSTMs is 2 and the number of heads for the multi-head attention region is 8. The number of predictive modes $K$ is set to 6. 

\subsection{Comparision Results}
\subsubsection{Comparison with State-of-the-art}


We evaluate the performance of the multimodal prediction model on the testing set and the validation set of the Argoverse dataset, setting the number of prediction modes $K$ to 6. As can be observed in Tab.~\ref{baselines1}, the performance of our model is optimal compared to the map-free methods. In particular, compared to the rencently-proposed map-free trajectory prediction method, CRAT-Pred\cite{schmidt2022crat}, it improves in the minADE, minFDE metrics by 0.13 and 0.31, respectively. Although our model does not use map information, it also presents competitive performance compared to the map-based methods TNT\cite{zhao2021tnt} (e.g, minADE = 0.94), PRIME\cite{song2022learning} (e.g, minADE = 1.22), MTPLA\cite{luo2020probabilistic} (e.g, minADE = 0.99), and holmes\cite{huang2020diversitygan} (e.g, minADE = 1.38).


The performance comparison results in the validation set are shown in Tab. \ref{baselines val}. As most trajectory prediction models are specifically designed to incorporate map information, they cannot be used to evaluate map-free predictions. However, there are some map-based models Tpcn\cite{ye_tpcn_2021}\textcolor{red}{$\dagger$}, LaneGCN\cite{liang_learning_2020}\textcolor{red}{$\dagger$}, HiVT-64\cite{zhou_hivt_2022}\textcolor{red}{$\dagger$}, which can exclude the map module and then be used for map-free prediction. It can be found that our model has a much better performance.

\subsubsection{Inference Speed}

Using the RTX 3090 GPU on the Argoverse validation set, our model was compared to multiple methods for inference speed. As shown in  Tab. \ref{inference speed}, all variants of our model have faster inference speeds than the baseline. For the model HiVT-64\cite{zhou_hivt_2022}\textcolor{red}{$\dagger$} with the map component removed, we not only improve in prediction accuracy, but also outperform in inference speed. Compared to the recently released map-free trajectory prediction model CRAT-Pred\cite{schmidt2022crat}, we have improved inference speed by 119 ms, and improved prediction performance metrics minADE and minFDE by 0.11 and 0.26 respectively. Compared to the HD map-based model mmTransformer\cite{liu_multimodal_2021}, our inference speed is twice as fast and comparable performance in trajectory prediction accuracy. These results show that our model is a fast inference and high prediction accuracy model for map-free trajectory prediction.
\begin{table}[h] 
\hspace{-0.1cm}
\renewcommand{\arraystretch}{1.7}
        \caption{The inference speed and the prediction performance of models on the Argoverse validation set.} 
	\scalebox{0.79}{
	\setlength{\tabcolsep}{2.5pt}{
        \begin{threeparttable}
	\begin{tabular}{l|l|c|ccc} 
		\toprule[1pt] 
		Model&Input&Speed(ms) $\downarrow$ &  minADE $\downarrow$ &  minFDE $\downarrow$&  MR $\downarrow$ \\
		\cmidrule(lr){1-6}
		HiVT-64\cite{zhou_hivt_2022}\textcolor{red}{$\dagger$} &Trajectory& 35  &0.77 &1.25 &14.00 \\
		CRAT-Pred\cite{schmidt2022crat} &Trajectory& 178  &0.85 &1.44& 0.17 \\
		mmTransformer\cite{liu_multimodal_2021} & Trajectory+Map&59& 0.72& 1.21& 9.20\\
		\cmidrule(lr){1-6}
            Ours(No Encoder) & Trajectory& 23  &0.89 & 1.52&19.50 \\
		Ours(No Spatial Interation) & Trajectory& 21  &0.88 & 1.61&21.40  \\
		Ours(No Temporal Interation) & Trajectory& 24  &0.76 & 1.24&13.90 \\
		\cmidrule(lr){1-6}
		Ours& Trajectory& 28  &0.74 & 1.18 & 11.69 \\
		\bottomrule[1pt] 
	\end{tabular}
        \end{threeparttable}
        \label{inference speed}
		} 
	}
 \vspace{-0.1cm}
\end{table}
\subsubsection{Qualitative Results}
\begin{figure}[H]
 	\centering
 	\includegraphics[width=0.49\textwidth]{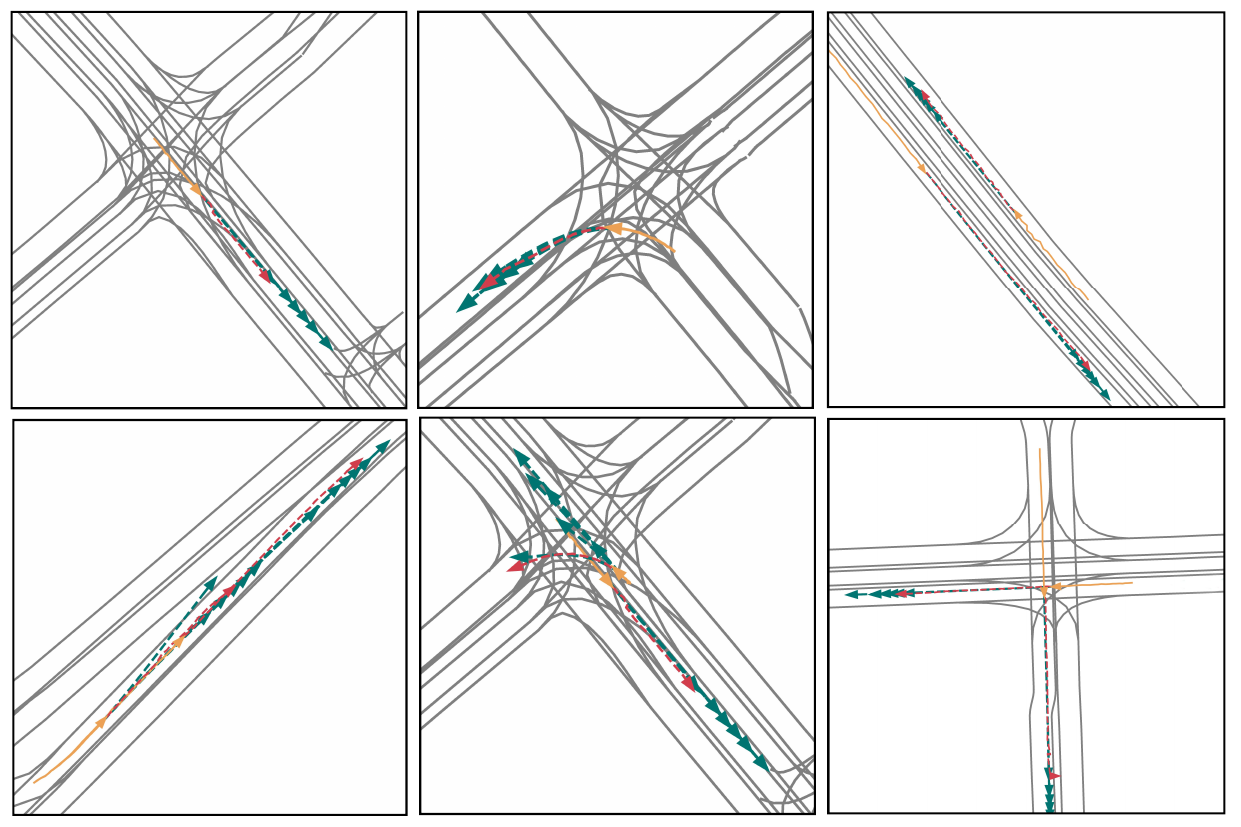}
 	\caption{Qualitative results. The past trajectories are shown in yellow, the ground-truth trajectories are shown in red, and the predicted trajectories are shown in green.}
 	\label{fig_result}
        \vspace{-0.3cm}
 \end{figure}

For clarity, we visualize the trajectory prediction results. As shown in Fig. \ref{fig_result}, our model can accurately, efficiently, and reasonably predict the behavior of agents in complex traffic scenarios with multiple modes. It can accurately infer various traffic behaviors such as turning, acceleration, merging, and straight driving.
 \subsection{Ablation Studies}
A series of ablation studies are carried out to analyse the effectiveness of single agent encoder, multiple agents interaction module and multidimensional attention mechanism.
\begin{table}[ht]
    \setlength\tabcolsep{2pt}
	\caption{Ablation experiments on the agoverse validation set} 
	\begin{center}
        \scalebox{0.68}{
        \setlength{\tabcolsep}{2pt}{
         \begin{tabular}{c|cccccccc|ccc} 
		\toprule[1pt]
            \multirow{4}*{Ablations}&\multicolumn{4}{c}{Encoder} & $\rightarrow$	&\multicolumn{2}{c}{Interaction}& ~& \multirow{4}*{minADE $\downarrow$} & \multirow{4}*{minFDE $\downarrow$}& \multirow{4}*{MR(\%) $\downarrow$}\\ 
            \cmidrule(lr){2-5} \cmidrule(lr){7-8}
            ~& Spatial &~& \multicolumn{2}{c}{Temporal}&~&\multirow{2.4}*{Spatial} & \multirow{2.4}*{Temporal}& &  ~&  ~& ~\\
            \cmidrule(lr){2-2} \cmidrule(lr){4-5}
            ~& SA & ~&LSTM&SA&~&~ & ~& &  ~&  ~& ~\\
		\cmidrule(lr){1-12}
		\multirow{4.4}*{Single Agent Encoder}&~&~ & \CheckmarkBold &\CheckmarkBold &~ & \CheckmarkBold & \CheckmarkBold& ~  &0.75 & 1.23&13.70\\
            ~&\CheckmarkBold &~&~ & \CheckmarkBold &~ & \CheckmarkBold & \CheckmarkBold& ~  &0.82 & 1.37&16.34 \\
		~&\CheckmarkBold &~&\CheckmarkBold & ~ &~ & \CheckmarkBold & \CheckmarkBold& ~  &0.76 & 1.24&13.93 \\
            ~&~ &~&\CheckmarkBold & ~ &~ & \CheckmarkBold & \CheckmarkBold& ~  &0.89 & 1.52&19.50 \\
            \cmidrule(lr){1-12}
		\multirow{2}*{Multiple Agents Interaction}&\CheckmarkBold &~&\CheckmarkBold & \CheckmarkBold & ~& ~ & \CheckmarkBold& ~  &0.88 & 1.61&21.40 \\
		~&\CheckmarkBold &~&\CheckmarkBold & \CheckmarkBold & ~&\CheckmarkBold & ~& ~  &0.76 & 1.24&14.10 \\
            \cmidrule(lr){1-12}
          \multirow{2}*{Temporal \& Spatial}&\CheckmarkBold &~&~ & ~ &~ & \CheckmarkBold & ~ & ~&1.18 & 1.97&29.00 \\
          ~&~ &~&\CheckmarkBold & \CheckmarkBold & ~ &~& \CheckmarkBold  & ~&0.91 & 1.67&22.75 \\
          \cmidrule(lr){1-12}
		All&\CheckmarkBold &~&\CheckmarkBold & \CheckmarkBold &~& \CheckmarkBold & \CheckmarkBold  & ~& \textbf{0.74} & \textbf{1.18}&\textbf{11.69} \\
		\bottomrule[1pt] 
	\end{tabular} 
        }
        }
	\end{center}
        \begin{tablenotes}
        \footnotesize
        \item SA: Self-Attention.
        \end{tablenotes}
	\label{Ablation experiments3} 
  \vspace{-0.2cm}
\end{table}
\subsubsection{Single Agent Encoder}
We investigated the impact of each component of the Single Agent Encoder module on the model performance individually, and the experimental results are shown in the Tab.~\ref{Ablation experiments3}. The comparative results show that our proposed multi-dimensional attention mechanism in temporal and spatial, when applied individually, can significantly improve the prediction accuracy of the model. When the multidimensional attention mechanism is applied to the module simultaneously, the minFDE improves by up to 0.15, indicating that multidimensional attention is one of the important factors in the performance improvement of the model. An interesting finding was that the performance of the model decreased significantly when we removed the LSTM component. This suggests that the LSTM plays an important role in capturing the temporal dynamics of individual agent trajectories.

\subsubsection{Multiple Agents Interaction}
Our proposed multiple agents interaction mechanism consists of the temporal interaction module and the spatial interaction module. To test the effects of these two modules, some ablation experiments are conducted and the results are shown in Tab.~\ref{Ablation experiments3}. We can observe that the model configured with both modules achieves higher performance (e.g, MR=11.69) than the model configured with either spatial interaction module (e.g, MR=14.10) or temporal interaction module (e.g, MR=21.40), proving both interaction modules are effective. We can also find an interesting result that the model configured with the spatial interaction module (e.g, minFDE = 1.61) exhibits higher performance than the model configured with the temporal interaction module (e.g, minFDE = 1.24), which demonstrates the multiple agent spatial relations modeling is more important.

\subsubsection{Temporal \& Spatial}

We conducted ablation experiments to assess the ability of the model to handle spatial-temporal state features, which is important for trajectory prediction models. As can be observed in Tab.~\ref{Ablation experiments3}, when the Encoder and Interaction modules, which handle spatial or temporal state features, are removed simultaneously, the minADE decreases by 0.17 and 0.44, respectively, demonstrating the effectiveness of handling spatial-temporal features. It can also be observed that the model with all temporal modules removed simultaneously (e.g, minADE = 1.18) has a greater impact on the performance of the model than the model with all spatial modules removed simultaneously (e.g, minADE = 0.91), indicating that the modeling of temporal state features is more significant for the map-free trajectory prediction model.

\section{Conclusion}
This paper presents an efficient trajectory prediction model that does not rely on maps. Advanced performance is achieved without the use of map information. It provides faster inference compared to those methods with comparable prediction performance. The core idea of the model is to encode single-agent's spatial-temporal information in the first stage and explore multi-agent spatial-temporal interactions in the second stage. By using a combination of attention mechanisms, LSTM, graph convolutional networks and temporal transformers in both phases, our model is able to learn the rich dynamics and interaction information of all agents. 








\bibliographystyle{IEEEtran}
\bibliography{IEEEexample}

\end{document}